# Learned Features are better for Ethnicity Classification

*Inzamam Anwar[1†], Naeem Ul Islam[2]*

Intelligent Systems Research Institute (ISRI), College of Information and Communication Engineering, Sungkyunkwan University, Suwon, South Korea
Emails: [1]inzimammanzoor@gmail.com   [2]naeem@skku.edu

***Abstract:*** *Ethnicity is a key demographic attribute of human beings and it plays a vital role in automatic facial recognition and have extensive real world applications such as Human Computer Interaction (HCI); demographic based classification; biometric based recognition; security and defense to name a few. In this paper we present a novel approach for extracting ethnicity from the facial images. The proposed method makes use of a pre trained Convolutional Neural Network (CNN) to extract the features and then Support Vector Machine (SVM) with linear kernel is used as a classifier. This technique uses translational invariant hierarchical features learned by the network, in contrast to previous works, which use hand crafted features such as Local Binary Pattern (LBP); Gabor etc. Thorough experiments are presented on ten different facial databases which strongly suggest that our approach is robust to different expressions and illuminations conditions. Here we consider ethnicity classification as a three class problem including Asian, African-American and Caucasian. Average classification accuracy over all databases is 98.28%, 99.66% and 99.05% for Asian, African-American and Caucasian respectively.*

† Corresponding author. All the codes are available for reproducing the results on request.

## 1. Introduction

Ethnicity identification is the process of recognizing the ethnic group of an individual from a face image. Human face provides wealth of information including identity, gender, age, race, expression etc. Among the demographic attributes of gender, age, ethnicity etc., ethnicity remains invariant through all lifetime and greatly supports face recognition systems, therefore, automatic facial ethnicity classification has been received increasing attention in recent years and several methods have been proposed. However, accurate and swift classification of different races based on human face in an uncontrolled environment is challenging. For efficient classification, one has to find race sensitive features from face images. These discriminative features can be differentiated in three categories [1] namely, chromatic/skin tone, local features, and global features. Due to similar skin colour for different races and extreme variation in illumination conditions for real world scenarios, skin tone alone cannot classify. However, combined with local or global descriptors classification accuracy can be boosted.

Feature based identification are based on feature extraction either globally or locally. In [2] a two class ethnicity classification for Asian and non-Asian images is presented. Rather than using full image, the method determine the confidence of different facial regions by employing modified Golden ration mask followed by using Support Vector Machine (SVM) classifier on facial features such as eyes, nose and mouth. Similarly, a two class ethnicity classification approach based on Linear Discriminant Analysis (LDA) [3] and Principle Component Analysis [4] are also presented. Nevertheless, these methods are limited and it is unclear about the efficiency of these approaches in more realistic conditions of multi-ethnic groups. The LDA approach in [3] was later enhanced with geometric features extracted using Gabor wavelet transform [5] for a three-class scenario.

Relationship between Iris texture and race group is studied by many researchers in past [1]. Results have shown that there is a strong relationship between iris texture and race [6-8]. But due to low resolution and distant images of subjects in real world, one cannot rely on cumbersome method of iris detection and identification. Lin et al. [9] have used Adaboost and Gabor filters to extract features which are classified using SVM and presented results on FERET database. Local Binary Pattern (LBP) is a non-parametric descriptor, which summarizes the local structures of images efficiently. In [10], demographic classification based on LBP is presented. In order to further improve the accuracy of race identification, various methods based on the fusion of different features are presented [11-13].

Recently deep learning methods have shown promising results in computer vision. With the breakthrough results in Large Scale Image Classification and Recognition Challenge (ILSVRC) [14] many researchers are using deep learning methods especially convolutional neural network (CNN) in different recognition and classification problems. These advancements in deep learning happen due to availability of more computational power and large datasets for training. Convolutional Neural Network consists of stack of layers which learn the translational invariance and hierarchal contextual features from input images. This type of deep learning have become popular and outperforms previous state of the art techniques in various domains [15-21].

This paper presents a framework to find ethnicity of the subject from face images using the deep convolutional neural network (CNN), pre-trained on large scale face database. This network is used as an independent feature extractor which are then classified with a linear classifier. Exhaustive experimentation is presented on ten databases. Results of our proposed approach strongly advocate that it is robust to different illumination conditions; expressions; accessories; background; age; and gender. The rest of the paper is organized in the following way: in Section 2, a brief introduction of CNN; VGG network; and pre-processing is presented. Section 3 provides introduction of datasets used, and presents results of in depth experimentation. In Section 4, conclusions from this study and experiments are drawn.

## 2. Proposed approach

### 2.1. Convolutional neural network

Convolutional Neural Network (CNN) are derived from basic neural networks. In neural networks, there are three types of layer, input, hidden and output. Each layer have several neurons stacked in it which takes input from the previous layers. In each layer, a neuron acts as a linear classifier. Each neuron performs some mathematical operation, usually computes a dot product of input with its respective weight, adds biasing, and applies a non-linearity.

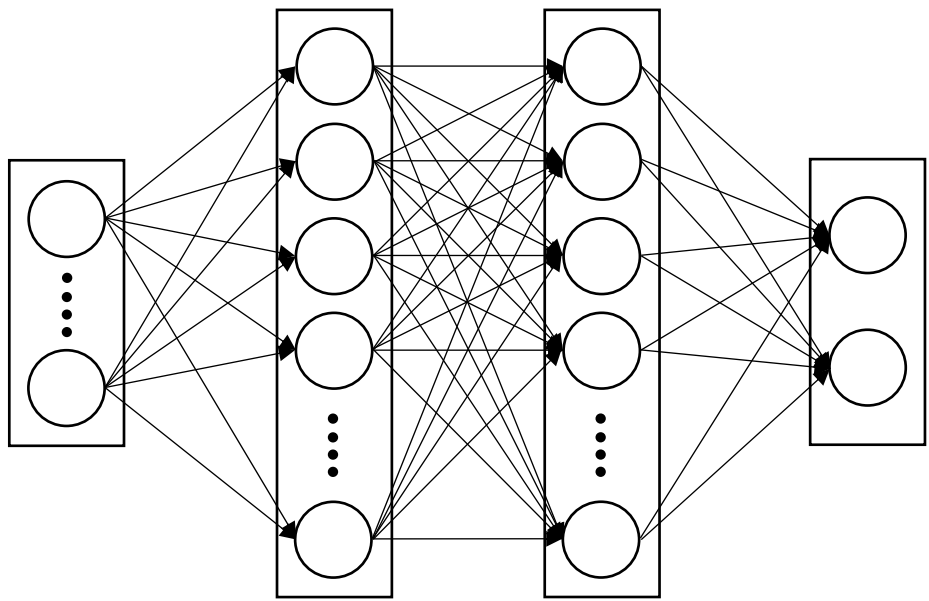

Fig. 1. A simple architecture of 3-layer fully connected neural network. Circles in each layer represent one neuron.

The output is fed into the neuron of next layer. Usually neurons between two adjacent layers are fully connected, and neurons within a single layer are not interconnected. Architecture of a 3-layer neural network is shown in figure 1. The problem with this simple neural network is scaling. For example if the input image has spatial size of 32x32, then one neuron in first hidden layer would have 32*32 = 1024 weights. But if the input size is increased to 200*200 then one neuron will have 4k weights and optimization of these much parameters will lead to over-fitting.

In CNN each neuron is connected to a small portion of input layer. The neuron computes a value by taking dot product between the weights and small input region. This layer is called convolutional layer (CONV layer). The size of the small region is called receptive field. The conv layer have neurons stacked in three dimensions, depth, height and width. The depth of output volume is defined by number of neurons. Height ($H$) and width ($W$) is defined by receptive filed or filter size, padding, and stride. The output volume can be calculated using the following formula given in (1).

$$W = H = \frac{N - F}{stride} + 1 \quad (1)$$

Where N is the input size and F is the filter size. Rectified Linear Unit (ReLU) layer is used to apply non-linearity (activation function) for each element of 3D volume obtained after convolution. An activation function should has a maxima and minima (monotonic), must be continuous and differentiable. Pooling layer comes after the ReLU layer which reduces the size of input volume in spatial domain. The reduction of spatial size is necessary to reduce the number of parameters in subsequent layers as number of neurons increases. This also helps to avoid overfitting and introduce translational invariance in the learned model. At the end of several conv+ReLU+pool layers, fully connected (FC) layers are used for classifying the input image based on learned features.

2.2. VGG face topology

For this study VGG-Face network is used. The architecture of this model is based on the Visual Geometry Group (VGG) deep convolutional neural network proposed in [22]. The main reason behind opting this network is that it is pre trained on a large face dataset of 2.6 million images. It presents comparable accuracy to state of the art networks, which are very deep than this, on benchmarks like Labelled Faces in Wild (LFW) and YouTube Faces Dataset [22].

VGG-Face net comprises of 37 layers which are divided in 6 blocks. In each convolutional layer (Conv Layer), filter size of 3*3 is used with stride 1 and padding 1. Frist two blocks have 2 convolutional layers each followed by ReLU layers. After 2[nd] ReLU layer there is max pool layer which reduces the spatial size of feature map to half. Next three block have three convolutional layers each, followed by ReLU layers and max pool layer after 3[rd] ReLU layer. Last block comprises of 3 fully connected layers followed by soft-max layer. Input image size for this architecture is 224*224*3 and it is trained on 2.6 M images with 2622 different classes (identities).

Trained network model is available for academic use [23]. The model is implemented in open source MatConvNet library [24].

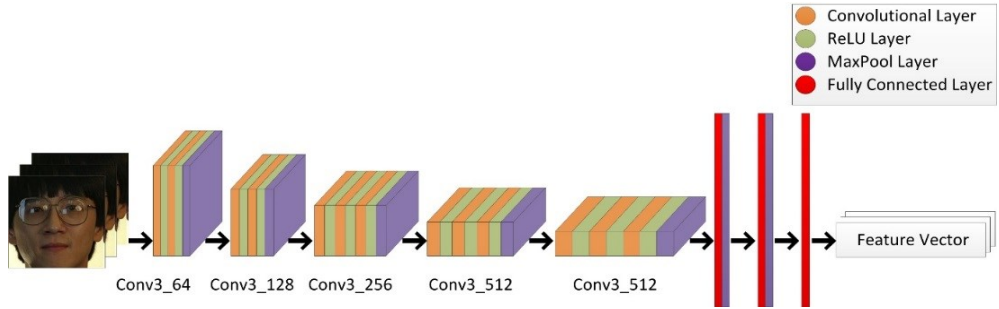

Fig. 2. VGG Face Network architecture. ConvX_Y, X represents the spatial size of the filter used and Y represents the number of filter used in particular convolutional layer. Feature vector is 4096 dimensional vector.

2.3. Support vector machine

SVM is a traditional binary classifier used in many different pattern classification applications. Other than pattern classification, it has also been successfully applied to bioinformatics; face detection and recognition; object detection and recognition; handwriting recognition; information and image retrieval; speaker and speech verification [25]. SVM have a good generalization properties. This is because it is based on the Structural Risk Minimization (SRM) principal which is rooted in the statistical learning theory. Supervised learning is used for training the machines with input data labelled with different classes. Principally, SVM finds a hyperplane, given by (2), which separates the two classes labelled with $\{-1,1\}$ with maximum margin. Margin is the distance between the two lines which pass through the data points which are closest to the opposite class. These data points are called Support Vectors (SV).

(2) $$f(x) = \sum_{i=1}^{n} w_i x_i + b$$

The line passing through the SV have equations $w \cdot x_i + b \geq 1$ for class +1 and $w \cdot x_i + b \leq -1$ for class -1.

(3) $$w \cdot x_1 + b = 1$$

(4) $$w \cdot x_2 + b = -1$$

$$w \cdot (x_1 - x_2) = 2$$

(5) $$\frac{w}{\|w\|} \cdot (x_1 - x_2) = \frac{2}{\|w\|}$$

Maximum margin can be found by minimizing (5), with subject to constraint, $y_i(w.x_i + b) \geq 1$. This form a Quadratic Programming (QP) problem. World data is usually not separable by the linear classifier. For this reason non-linear kernel functions are used to map data from input space to a higher dimensional space where data can be linear separable.

2.4. Pre-processing

All the face images before testing and training are aligned using the dense fiducial points of eyes, eye brows, nose and mouth, in total 68 points. For fiducial point detection and alignment, Dlib library [26] is used with its python wrapper. Histogram of Oriented Gradients (HOG), image pyramid and sliding window technique is used to find the bounding box for face and [27] is used for fiducial point detection. Images are then align to one common reference face by finding the minimum sum of squared difference between these fiducial points. This process is done by Ordinary Procrustes Analysis, given in (6).

$$(6) \quad \sum_{i=1}^{68} \left\| y_i^t - (Rx_i^t + T) \right\|^2$$

$R$ is the rotational 2x2 matrix, $T$ is a translational 2x1 vector and $y_i$ and $x_i$ are the coordinates of fiducial points of reference and target images respectively. For cropping the faces, Eye detection is performed using Haar cascaded classifier implemented in Matlab.

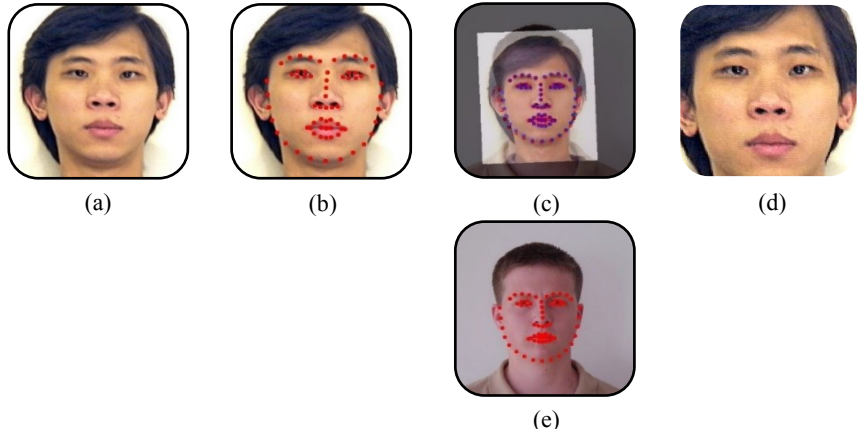

Fig. 3. Face alignment and crop is represented. (a) is the input image, (b) shows the 68 landmarks detected, (c) represents the layover of reference fiducial points over input image, (d) face cropped and (e) reference image.

A 224*224 patch is cropped with respect to line connecting the eye centres such that for every face it should roughly come in the same row. Mean of the data used to train is subtracted from the input images. VGG-Face Net is used to extract the 4096-d feature vector. Support vector machine (SVM) with linear kernel is used for training, on the feature vectors obtained from convolutional neural network (CNN).

## 2.5. Visualization

Figure 4 shows the 2D t-SNE (t-Distributed Stochastic Neighbor Embedding) [28] plot of the whole database and the features extracted from VGG Face network. The spatial distance indicates the similarity between the images and show strong clustering of images from same class. Although the feature vector, which is 4096-dimensional, is shown in 2D space it still shows superiority of network in increasing inter-class separability and intra-class compactness.

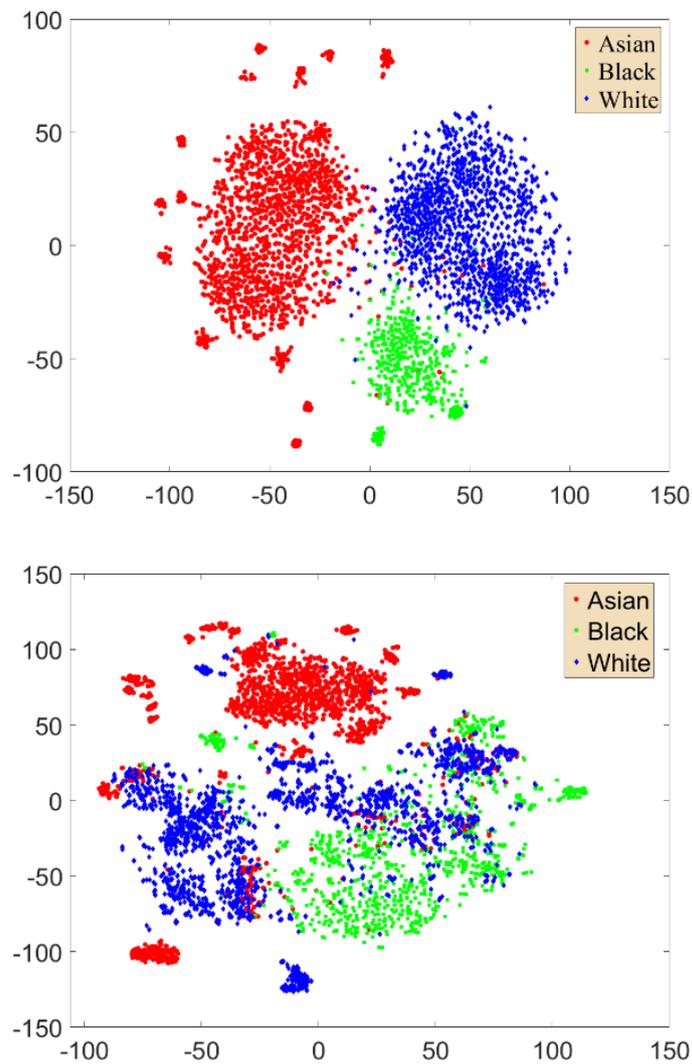

Fig. 4. 2-D t-SNE plot of features obtained from CNN (above) and complete dataset (below) is shown. These plots are generated with 1K iterations and perplexity 30. The face images are resized to 38*38.

## 3. Experiments and results

### 3.1. Databases

The proposed process is evaluated on ten different databases including Computer Vision Lab (CVL) face database [29], Chicago Face Database (CFD) [30], Face Recognition Technology (FERET) database [31] , Multi-racial mega resolution (MR2) face database [32], UT Dallas face database [33], Psychological Image Collection at Stirling (PICS) Aberdeen [34], Japanese Female Facial Expression (JAFFE) database [35], CAS-PEAL-R1 [36], Montreal Set of Facial Displays of Emotion Database (MSFDE) [37] and Chinese University of Hong Kong face database (CUFC) [38]. Purpose of using different databases is to evaluate the proposed methodology rigorously and also these datasets are publically (without any cost) available. These databases have variations in pose and viewpoints; illumination condition; different expressions; and emotions. Detailed description of the datasets are provided in table 1.

Table 1. Detailed breakdown of datasets is presented. N/A means that database has no images for that class, N (normal expression), Var-ill (variable illumination), S (smile with teeth visible (T) and invisible (NT)), A (angry), F (fear), HC (Happy mouth closed), HO (Happy mouth open), Acc. (accessories), Age. (Aging), Bac. (Background), Dist. (different viewpoint), Exp. (expression).

| Database | Race Asian | | | Black | | | White | | |
|---|---|---|---|---|---|---|---|---|---|
| MR2 | 20 | | | 32 | | | 22 | | |
| UT | 6 | | | 158 | | | 582 | | |
| MSDFE | 175 | | | 115 | | | 176 | | |
| CUFC | 188 | | | N/A | | | N/A | | |
| JAFFE | N 10 | Var-ill-exp 213 | | N/A | | | N/A | | |
| Aberdeen | 19 | | | N/A | | | N 86 | Var-ill 633 | |
| FERET | fa 130 | fb 124 | r 156 | fa 75 | fb 70 | r 64 | fa 605 | fb 602 | r 562 |
| CVL | N/A | | | N/A | | | N 113 | S/NT 113 | S/T 113 |
| CFD | 109 | | | N A F HC HO | 197 81 82 78 83 | | N A F HC HO | 184 71 65 73 68 | |
| CASPEALR1 | Acc. Age. Bac. Dist. Exp. N Lighting | 1859 66 640 318 1879 1038 1341 | | N/A | | | N/A | | |
| Total | 8291 | | | 1035 | | | 4068 | | |

### 3.2. Experiments

The results presented here are obtained using ten-fold cross validation. For each fold, nine databases are used to train linear SVM with five-fold cross validation and remaining one database is used for testing. In all cases, test set and train set are mutually exclusive meaning image of an individual in test is not in the train set, in any case, or vice versa. Although, train or test set may contain more than one image per individual. As most of the databases have different poses, illumination and lighting conditions, neutral face images are only used for training. However, all images are used for testing. This shows the robustness of our technique. Oversampling is used if the number of images for one class is less than others.

Figure 5 and 6 shows the results for databases which have at least two classes. In Aberdeen database there are only two classes Asian and White. Accuracy of Asian class in CFD database is low because there is strong visual resemblance of misclassified persons with Black class. For FERET database separate confusion matrices are shown for three types of sets named 'fa', 'fb' and 'r'. The first set 'fa', contains images of subjects with frontal pose, second set 'fb' contains images with frontal pose with different expressions and third set, 'r' contains images with a pose, head turned by ±15 degrees.

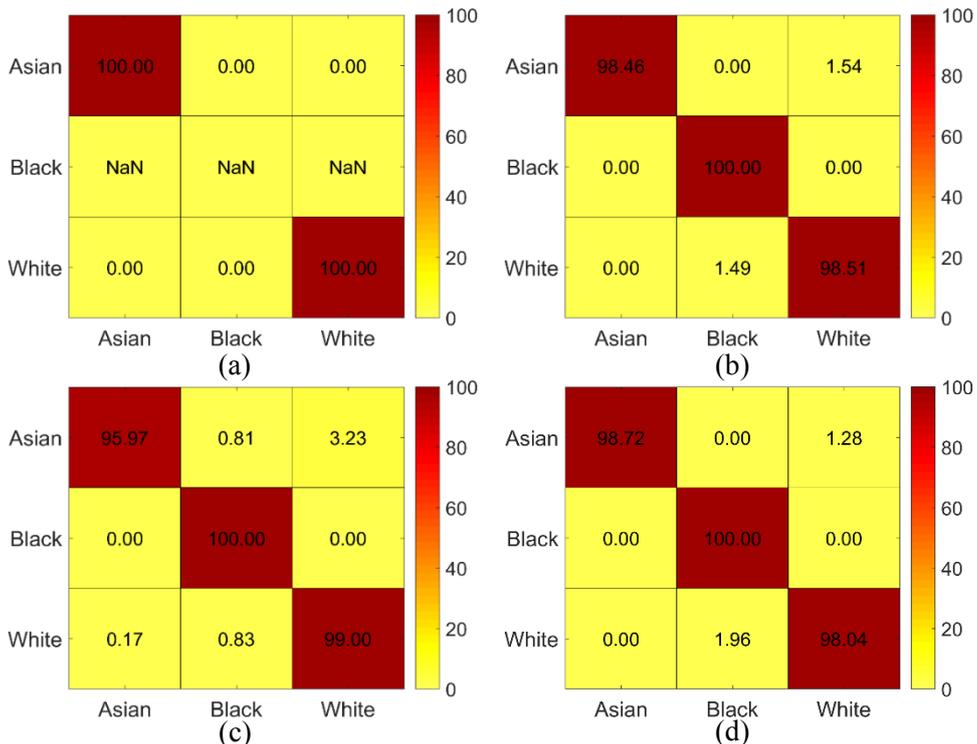

Fig. 5. Confusion matrices for Aberdeen (a), FERET fa (b), FERET fb (c), and FERET r (d) are shown respectively.

Classification results for MR2, MSDFE, UT Dallas and CFD database with neutral expressions is shown in figure 6. In CFD, for some subjects from Black and

White class, there are images with expression. These expression includes angry, fear, smile with teeth visible and smile with teeth not visible. Results for these images are shown in figure 7.

For CASPEAL database, results are shown in figure 8. It includes frontal images with different characteristics; including normal expression; variation in distance between camera and person; wearing different accessories (Sunglasses and Hat); aging (photos of subjects taken with an interval of 6 months); with different illumination conditions; different backgrounds; and different expressions (open mouth, frown, closed eyes, smile and surprise). In case of different illumination conditions the accuracy is low due to extreme changes in illumination. The illumination condition are represented in fig 3 of [36]. It is observed that the illumination condition in which fluorescent light is below the face of the target and where some part of the face is not illuminated leads to erroneous results.

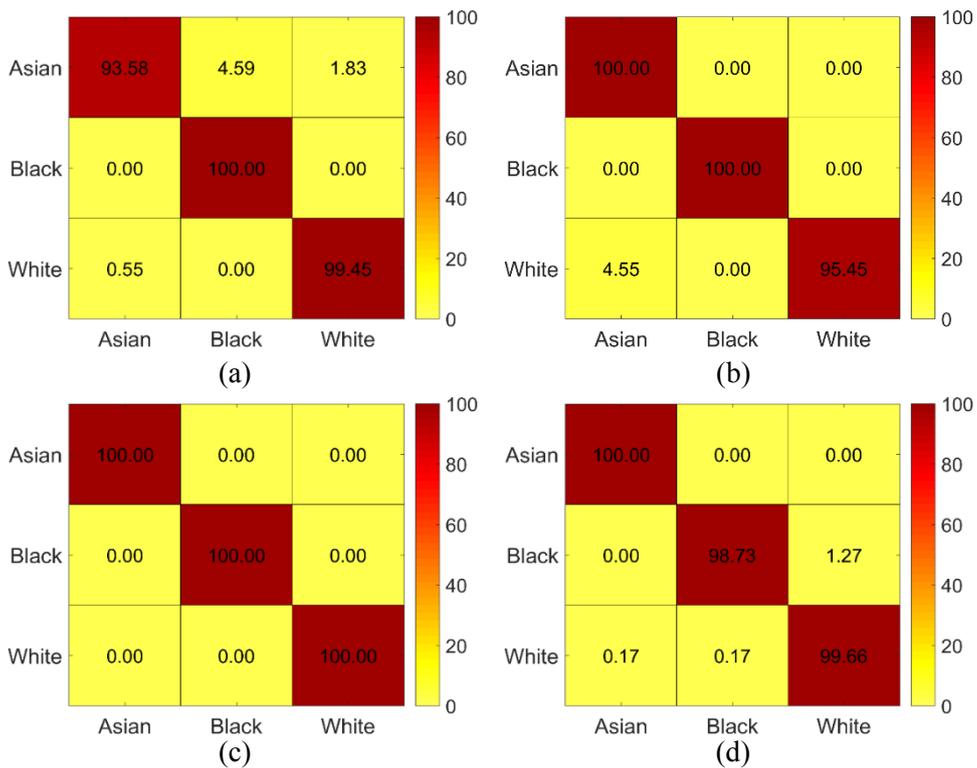

Fig. 6. Confusion matrices for CFD (with no expression) (a), MR2 (b), MSDFE (c) and UT Dallas (d) Datasets.

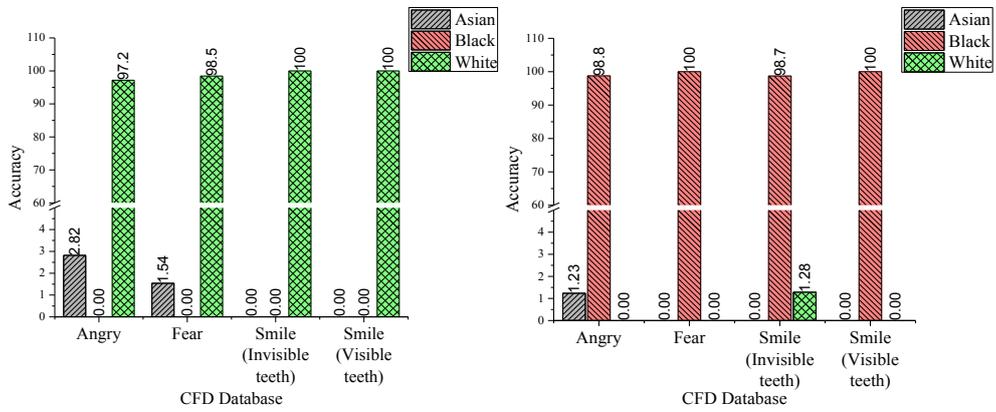

Fig. 7. Results for CFD Database for White and Black subjects which have different expressions.

These results in addition to expression results presented with CASPEAL database strongly suggest that this approach is robust to commonly encountered expressions.

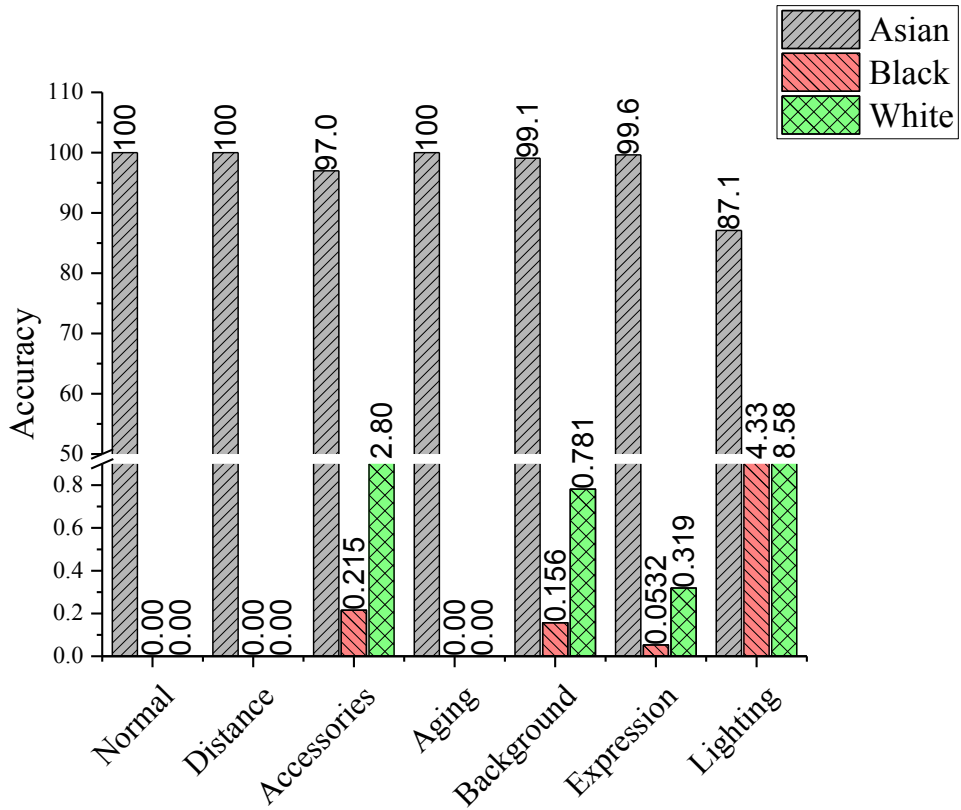

Fig. 8. Results for CASPEAL R1 dataset.

3.3. Comparison

In table 2, comparison between our findings and previous best results in literature are shown. Most of these results are presented on FERET and CASPEAL face database. The results show that our approach performs better than all previous methods. In [2] race classification is taken as two class (Asian/Non-Asian) problem, which is an easy problem then our 3-class classification. They have used images for test and train from same pool of images which does not include cross database problems. In [8], they also used training and testing images from same database. Although the accuracy in [12] is comparable to our method but the training and test set are not explicitly defined and it is difficult to say whether the images in training and test set are mutually exclusive or not.

Table 2. Comparison with existing methods in literature. *represents the average accuracy over fa, fb and r subsets of FERET database. ** shows the average accuracy over all the datasets.

| Method | Accuracy | | | Database |
|---|---|---|---|---|
| | Asian (%) | Black (%) | White (%) | |
| Manesh et al.[2] | 96 | N/A | N/A | FERET |
| Gutta et al. [8] | 92 | 92 | 92 | FERET |
| Muhammad et al. [12] | 99.47 | 98.99 | 100 | FERET |
| Wu et al. [39] | 95 | 96.1 | 93.5 | FERET |
| Roomi et al. [40] | 79.13 | 90 | 90.9 | FERET+Yale |
| Ou et al. [41] | 83.5 | N/A | N/A | FERET |
| **Ours** | **98.46** | **100** | **100** | **FERET (fa)** |
| **Ours** | **95.97** | **100** | **99** | **FERET (fb)** |
| **Ours** | **98.72** | **100** | **98.04** | **FERET (r)** |
| **Ours*** | **97.72** | **100** | **99.01** | **FERET** |
| **Ours**** | **98.28** | **99.66** | **99.05** | **All 10** |

4. Conclusion

This paper presents an approach based on Convolutional Neural Network (CNN) to estimate the ethnicity of an individual from facial images. CNN is used here as a standalone feature extractor and linear Support Vector Machine (SVM) is used to classify. Ten different datasets are used for exhaustive experimentation and results presented here outperforms the previous techniques. The experiments also indicate that our approach is robust to different illumination conditions; backgrounds; age; gender; and various expressions. This work finds its application in Human Computer Interaction (HCI); face recognition; biometric based recognition; and surveillance and defence. Future work may include the development of a framework which can efficiently classify sub-ethnic group e.g. classifying Koreans from Chinese or Europeans from Americans.

# References

1. F u , S . , H . H e , Z . G . H o u . Learning race from face: A survey. – IEEE Transactions on Pattern Analysis and Machine Intelligence, Vol. **36**, 2014, No. 12, pp. 2483-2509.
2. M a n e s h , F . S . , M . G h a h r a m a n i , Y . P . T a n . Facial part displacement effect on template-based gender and ethnicity classification. – In: Proc. of 11th International conference on Control Automation Robotics & Vision (ICARCV), Singapore, 2010, pp. 1644-1649.
3. X i a o g u a n g , L u . , A . K . J a i n . Ethnicity identification from face images. – In: Proc. of SPIE International Symposium on Defense and Security: Biometric Technology for Human Identification, Orlando, Florida, USA, 2004, pp.114-123.
4. T i n , H . H . , M . M . S e i n . Race identification for face images. – ACEEE International Journal on Information Technology, Vol. **1**, 2011, No. 02, pp. 118-120.
5. D u a n , X . D . , C . R . W a n g , X . D . L i u , Z . J . L i , J . W u , H . L . Z h a n g . Ethnic features extraction and recognition of human faces. – In: Proc. of the 2nd International Conference on Advanced Computer Control (ICACC), Shenyang, Liaoning, China, 2010, pp.125-130.
6. Z h a n g , H . , Z . S u n , T . T a n , J . W a n g . Ethnic classification based on iris images. – In: Proc. of Chinese conference on Biometric Recognition (CCBR), Beijing, China, 2011, pp. 82-90.
7. X i e , Y . , K . L u u , M . S a v v i d e s . A robust approach to facial ethnicity classification on large scale face databases. – In: Proc. of 5th IEEE International conference on Biometrics: Theory, Applications and Systems (BTAS), Arlington, VA, USA, 2012, pp. 143-149.
8. G u t t a , S . , J . R . H u a n g , P . J o n a t h o n , H . W e c h s l e r . Mixture of experts for classification of gender, ethnic origin, and pose of human faces. – IEEE Transactions on Neural Networks, Vol. **11**, 2000, No. 4, pp. 948-60.
9. L i n , H . , H . L u , L . Z h a n g . A new automatic recognition system of gender, age and ethnicity. – In: Proc. of the 6th World Congress on Intelligent Control and Automation (WCICA), Dalian, China, 2006, Vol. 2, pp. 9988-9991.
10. Y a n g , Z , H . A i . Demographic classification with local binary patterns. – In: Proc. of the International Conference on Biometrics (ICB), Seoul, South Korea, 2007, pp. 464-473.
11. W a d y , S . H . , H . O . A h m e d . Ethnicity Identification based on Fusion Strategy of Local and Global Features Extraction. – International Journal of Multidisciplinary and Current research, Vol. **4**, 2016, No. 2, pp. 200-205.
12. M u h a m m a d , G , M . H u s s a i n , F . A l e n e z y , G . B e b i s , A . M . M i r z a , H . A b o a l s a m h . Race classification from face images using local descriptors. – International Journal on Artificial Intelligence Tools, Vol. **21**, 2012, No. 5, pp. 1250019 (24 pages).
13. S a l a h , S . H . , H . D u , N . A l - J a w a d . Fusing local binary patterns with wavelet features for ethnicity identification. – In: Proc. Of World Academy of Science, Engineering and Technology (WASET), Çanakkale, Turkey, 2013, No. 79, pp. 471-477.
14. K r i z h e v s k y , A . , I . S u t s k e v e r , G . E . H i n t o n . Imagenet classification with deep convolutional neural networks. – In: Proc. of Advances in Neural Information Processing Systems (NIPS), Nevada, USA, 2012, pp. 1097-1105.
15. T a i g m a n , Y . , M . Y a n g , M . A . R a n z a t o , L . W o l f . Deepface: Closing the gap to human-level performance in face verification. – In: Proc. of IEEE conference on Computer Vision and Pattern Recognition (CVPR), Columbus, OH, USA, 2014, pp. 1701-1708.
16. K i m , Y . Convolutional neural networks for sentence classification. arXiv preprint arXiv:1408.5882. 2014 Aug 25.
17. G i r s h i c k , R . , J . D o n a h u e , T . D a r r e l l , J . M a l i k . Rich feature hierarchies for accurate object detection and semantic segmentation. – In: Proc. of IEEE conference on Computer Vision and Pattern Recognition, Columbus, OH, USA, 2014, pp. 580-587.
18. K a r p a t h y , A . , G . T o d e r i c i , S . S h e t t y , T . L e u n g , R . S u k t h a n k a r , L . F e i - F e i . Large-scale video classification with convolutional neural networks. – In: Proc. of IEEE conference on Computer Vision and Pattern Recognition (CVPR), Columbus, OH, USA, 2014, pp. 1725-1732.

19. Ji, S., W. Xu, M. Yang, K. Yu. 3D convolutional neural networks for human action recognition. – IEEE Transactions on Pattern Analysis and Machine Intelligence, Vol. **35**, 2013, No. 1, pp. 221-231.
20. Levi, G., T. Hassner. Age and gender classification using convolutional neural networks. – In: Proc. of IEEE conference on Computer Vision and Pattern Recognition (CVPR) Workshops, Boston, USA, 2015, pp. 34-42.
21. Pinheiro, P. H., R. Collobert. Recurrent Convolutional Neural Networks for Scene Labeling. – In: Proc. of International conference on Machine Learning (ICML), Beijing, China, 2014, pp. 82-90.
22. Parkhi, O. M., A. Vedaldi, A. Zisserman. Deep Face Recognition. – In: Proc. of British Machine Vision Conference (BMVC), Swansea, UK, 2015, Vol. **1**, No. 3, p. 6.
23. http://www.robots.ox.ac.uk/~vgg/software/vgg_face/
24. Vedaldi, A., K. Lenc. Matconvnet: Convolutional neural networks for Matlab. – In: Proc. of 23rd ACM International Conference on Multimedia, Brisbane, Australia, 2015, pp. 689-692.
25. Hearst, M. A., S. T. Dumais, E. Osuna, J. Platt, B. Scholkopf. Support vector machines. – IEEE Intelligent Systems and their Applications, Vol. **13**, 1998, No. 4, pp. 18-28.
26. King, D. E. Dlib-ml: A machine learning toolkit. – Journal of Machine Learning Research, Vol. **10**, 2009, pp.1755-1758.
27. Kazemi, V., J. Sullivan. One millisecond face alignment with an ensemble of regression trees. – In: Proc. of IEEE conference on Computer Vision and Pattern Recognition (CVPR), Columbus, OH, USA, 2014, pp. 1867-1874.
28. Maaten, L. V., G. Hinton. Visualizing data using t-SNE. – Journal of Machine Learning Research. Vol. **9**, 2008, pp. 2579-2605.
29. Franc, S., P. Peer, B. Batagelj, S. Juvan, J. Kovac. Color-based face detection in the '15 seconds of fame' art installation. – In: Proc. of conference on Computer Vision / Computer Graphics Collaboration for Model-based Imaging Rendering, image Analysis and Graphical special Effects (MIRAGE), Versailles, France, 2003, pp. 38-47.
30. Ma, D. S., J. Correll, B. Wittenbrink. The Chicago face database: A free stimulus set of faces and norming data. – Behavior research methods, Vol. **47**, 2015, No. 4, pp. 1122-1135.
31. Phillips, P. J., H. Moon, S. A. Rizvi, P. J. Rauss. The FERET evaluation methodology for face-recognition algorithms. – IEEE Transactions on Pattern Analysis and Machine Intelligence, Vol. **22**, 2000, No. 10, pp. 1090-1104.
32. Strohminger, N., K. Gray, V. Chituc, J. Heffner, C. Schein, T. B. Heagins. The MR2: A multi-racial, mega-resolution database of facial stimuli. – Behavior Research Methods, Vol. **48**, 2016, No. 3, pp. 1197-1204.
33. Minear, M., D. C. Park. A lifespan database of adult facial stimuli. – Behavior Research Methods, Instruments, & Computers, Vol. **36**, 2004, No. 4, pp. 630-633.
34. Hancock, P. Psychological image collection at stirling (pics). Web address: http://pics.psych.stir.ac.uk. 2008 Mar.
35. Lyons, M. J., J. Budynek, S. Akamatsu. Automatic classification of single facial images. – IEEE Transactions on Pattern Analysis and Machine Intelligence, Vol. **21**, 1999, No. 12, pp. 1357-1362.
36. Gao, W., B. Cao, S. Shan, X. Chen, D. Zhou, X. Zhang, D. Zhao. The CAS-PEAL large-scale Chinese face database and baseline evaluations. – IEEE Transactions on Systems, Man, and Cybernetics-Part A: Systems and Humans, Vol. **38**, 2008, No. 1, pp. 149-161.
37. Beaupré, M. G., N. Cheung, U. Hess. The Montreal set of facial displays of emotion [Slides]. Available from Ursula Hess, Department of Psychology, University of Quebec at Montreal, PO Box. 2000, 8888. Web address: http://psychophysiolab.com/en/accueil.php.
38. Wang, X., X. Tang. Face photo-sketch synthesis and recognition. – IEEE Transactions on Pattern Analysis and Machine Intelligence, Vol. **31**, 2009, No. 11, pp. 1955-1967.

39. W u , B . , H . A i , C . H u a n g . Facial image retrieval based on demographic classification. – In: Proc. of 17th International conference on Pattern Recognition (ICPR), Cambridge, England, UK, 2004, Vol. 3, pp. 914-917.
40. R o o m i , S . M . , S . L . V i r a s u n d a r i i , S . S e l v a m e g a l a , S . J e e v a n a n d h a m , D . H a r i h a r a s u d h a n . Race classification based on facial features. – In: Proc. of 3rd National conference on Computer Vision, Pattern Recognition, Image Processing and Graphics (NCVPRIPG), Hubli, India, 2011, pp. 54-57.
41. O u , Y . , X . W u , H . Q i a n , Y . X u . A real time race classification system. – In: Proc. of IEEE International conference on Information Acquisition (ICIA), Hong Kong, China, 2005, pp. 378-383.